\documentclass{article} 
\usepackage{iclr2020_conference,times}

\usepackage{amsmath,amsfonts,bm}

\def\eqref#1{equation~\ref{#1}}

\def\1{\bm{1}}

\DeclareMathAlphabet{\mathsfit}{\encodingdefault}{\sfdefault}{m}{sl}
\SetMathAlphabet{\mathsfit}{bold}{\encodingdefault}{\sfdefault}{bx}{n}

\usepackage{hyperref}
\usepackage{url}
\usepackage[pdftex]{graphicx}
\usepackage{tabularx}
\usepackage{float}

\title{How not to give a FLOP \large \\ Combining regularization and pruning for efficient inference}

\author{Tai Vu, Emily Wen, Roy Nehoran \\
Department of Computer Science\\
Stanford University\\
\texttt{\{ taivu, ewen22, royn \}@stanford.edu} \\
}

\iclrfinalcopy 
\begin{document}

\maketitle

\begin{abstract}
The challenge of speeding up deep learning models during the deployment phase has been a large, expensive bottleneck in the modern tech industry. In this paper, we examine the use of both regularization and pruning for reduced computational complexity and more efficient inference in Deep Neural Networks (DNNs). In particular, we apply mixup and cutout regularizations and soft filter pruning to the ResNet architecture, focusing on minimizing floating-point operations (FLOPs). Furthermore, by using regularization in conjunction with network pruning, we show that such a combination makes a substantial improvement over each of the two techniques individually.
\end{abstract}

\section{Introduction}

Deep Convolutional Neural Networks (CNNs) have achieved great results and become one of the leading methods for sophisticated tasks, such as computer vision and natural language processing, in recent years \citep{krizhevsky2012imagenet,simonyan2014very}. However, they are also very computationally expensive, which has hindered their implementation in devices with limited hardware resources. This is largely due to the neural networks being over-parameterized and having significant amounts of redundant information stored in their layers \citep{tzelepis2019deep}. Furthermore, these computationally expensive DNNs consume a great deal of energy, and thus also be fiscally expensive to use. In order for DNNs to be widely utilized, it is important for their computational inefficiency to be alleviated without sacrificing accuracy. This not only reduces the amount of hardware resources needed, but also the energy and cost required to deploy them.

One popular technique harnessed in previous work for boosting model performance is regularization, which has been widely studied in the deep learning domain. Several common regularization approaches are parameter norm penalty, early stopping, data augmentation and the addition of noise during training time. An example of this is mixup, a data augmentation method that adds convex combinations of pairs of training instances and their labels \citep{mixup}. By training neural networks on such combinations of inputs, mixup improves the generalization errors of cutting-edge network architectures, alleviates the memorization of corrupt labels, and increases the sensitivity to adversarial examples. Likewise, the cutout approach in \citet{cutout} randomly masks out contiguous sections of the input images to CNNs, thereby improving robustness and yielding higher accuracy level \citep{cutout}. Alternatively, the technique dropout drops, or temporarily removes, random units in a neural network at intermediate layers \citep{dropout}. However, previous works have emphasized the use of regularization for preventing overfitting and guaranteeing that the resulting models generalize well to unseen data. They seemed to overlook the computational complexity of such regularized models in the inference process. Particularly, few papers have tested whether regularization speeds up the calculations of CNNs or introduces computational overhead during inference time.

Another feasible measure for high-performance model inference is network pruning, which aims to reduce the number of parameters. Recent research has harnessed magnitude-based pruning by eliminating all weights of low magnitude and retraining the networks to fine tune the remaining weights \citep{han2015,han2016}. This approach has been shown to significantly compress the size of common CNN architectures like AlexNet and VGG-16 with minimal loss of predictive accuracy. Nevertheless, most papers have focused on limiting the memory usage of deep learning models and embedding them on mobile devices. There is little research on using network pruning for reducing floating-point calculations, improving computational processes, and minimizing energy consumption.  Furthermore, most works have not examined the combination of pruning and advanced regularization methods like mixup and cutout to attain better performance. Since mobile phones also have limited processing powers, looking into this problem could improve the ability to run complex neural networks on them in the future.

In this paper, we examine the use of both regularization and pruning for reduced computational complexity and more efficient inference in Deep Neural Networks (DNNs). In particular, we apply two regularization methods called mixup and cutout and a pruning technique named soft filter pruning to the ResNet architecture, focusing on minimizing floating-point operations (FLOPs). Furthermore, by utilizing regularization in conjunction with pruning unimportant convolutional filters below some threshold in the network, we show that the combination contributes to a marked improvement over either of the two methods individually.

Our main focus in this paper is looking into whether pruning methods and regularization techniques could work in conjunction, rather than interfere with each other, to reduce computational inefficiency and keep accuracy rates comparable. On top of that, we determine the impact of applying regularization for the purpose of improving efficiency, a context that has been relatively unexplored. Along with this, we demonstrate that combining the methods produces much higher accuracies than just pruning. Overall, through this paper, we innovate a method that allows ResNet, and potentially other similar architectures, to be used in a less computationally expensive fashion. This more efficient implementation also results in important reductions in the energy, and thus cost, to deploy DNNs.

\section{Related Work}

Deep Neural Networks (DNNs) have achieved great outcomes and become one of the cutting-edge technologies for sophisticated tasks such as computer vision, natural language processing, and speech recognition. However, they are currently difficult to implement for mobile applications and embedded systems because they require a huge amount of storage. As a result, creating DNNs that are smaller and more efficient is very useful to make them more widely accessible for use within mobile applications.

\subsection{Pruning}
Some research has illustrated that there is significant redundancy in the parameterization of many neural networks \citep{denil2013predicting}, which consumes memory resources and slows down computation speeds. Thus, many research papers have emphasized reducing the number of parameters as a feasible way of compressing deep learning models and reducing their complexity.

Following that approach, early works like \citet{lecun1990optimal} and \citet{hassibi1993second} introduced weight pruning using the second derivative of the loss function with respect to the parameters to compute their saliencies. Instead, recent attempts, such as \citet{han2015} and \citet{han2015learning}, implemented magnitude-based network pruning, in which they remove all weights below some threshold and retrain the remaining networks to fine tune the weights for sparse connections. This practice sharply reduces the numbers of parameters in some popular CNN architectures by $9\times$ to $13\times$ without a significant loss of predictive accuracy. Likewise, \citet{collins2014memory} examined the application of sparsity-inducing regularizers to the training process of Convolution Neural Networks (CNNs), and this method reduces the memory usage of AlexNet by $4\times$. Furthermore, the recently developed architecture SqueezeNet attains even more success by compressing AlexNet by $50\times$ with a comparable accuracy level on the ImageNet dataset \citep{iandola2016squeezenet}.

While limiting the quantity of parameters proves to be a viable approach, most recent studies have been focused on popular network architectures including fully connected networks and CNNs. There is little research on less familiar structures like RNNs, LSTM, GANs, and so on. Additionally, many research papers have tested their models on classic datasets like MNIST and ImageNet and simple CNNs like LeNet or AlexNet, with little consideration of other datasets and more advanced architectures such as ResNet, GoogleNet, UNet, DenseNet, etc. 

While pruning parameters can reduce the sizes of the models, thereby minimizing memory usage and energy consumption, they do not necessarily increase computation speeds. In fact, the majority of the parameters removed are from the fully connected layers where the computation costs are low, whereas convolutional operations tend to dominate computation complexity \citep{li2016pruning}. Furthermore, high-performance inference on pruned models that contain sparse connections typically requires purpose-built hardware for performing sparse matrix-vector operations. The representation of sparse matrices also results in additional storage overheads, which raises the network’s net memory footprint \citep{zhu2017prune}.

\subsection{Regularization}
Because convolutional neural networks tackle complex learning problems, they are large and thus susceptible to overfitting. Regularization techniques can mitigate this issue and thus boost accuracy rates. Moreover, this practice can counteract the decrease in accuracy resulting from using pruning. Some common regularization  approaches are parameter norm penalty, early stopping, data augmentation, and the addition of noise during training time.

Dropout is a common regularization technique utilized for fully-connected layers. It is implemented by dropping, or temporarily removing, random units in a neural network \citep{dropout}. This is comparable to adding noise to a model's hidden units. While dropout removes parameters at intermediate layers, the technique cutout randomly occludes sections of the input images \citep{cutout}. By doing so, the model is able to focus more on less prominent features of the image rather than important features, i.e. those with high activations in intermediate layers of the network. This practice improves robustness and yields higher accuracy levels. Another variation of regularization, mixup, applies data augmentation \citep{mixup}. It enlarges the training dataset by taking the linear interpolations of two examples and their labels from the training data. By including these constructed examples in training, the neural network becomes more robust against adversarial examples and lessens the memorization of corrupt labels. Overall, mixup improves the generalization errors of these models on popular datasets.

Additionally, Structured Sparsity Learning (SSL) combines structural regularization and locality optimization to improve accuracy on large networks \citep{ssl}. It employs group Lasso regularization to alter the structure of the filters, channels, filter shapes, and depth structures within DNNs to construct compressed networks. By doing so, it directly learns a more compact representation of larger DNNs and reduces computational costs.

Previous works have emphasized the use of regularization for preventing overfitting and guaranteeing that the resulting models generalize well to unseen data. Furthermore, the impact of combining pruning methods and regularization techniques has not been thoroughly explored by previous studies, but will be scrutinized in this paper.

\section{Methods}

\subsection{Regularization}
We harnessed mixup, a data augmentation method that adds convex combinations of pairs of examples and their labels to the training data \citep{mixup}. Specifically, this technique uses new training examples of the form:
\begin{equation*}
\begin{aligned}
\tilde x = \lambda x_i + (1 - \lambda) x_j, & \quad \text{ where } x_i, x_j \text{ are input vectors.} \\
\tilde y = \lambda y_i + (1 - \lambda) y_j, & \quad \text{ where } y_i, y_j \text{ are one-hot label encodings.} 
\end{aligned} 
\end{equation*}
Here, $(x_i, y_i)$ and $(x_j, y_j)$ are two instances randomly drawn from the training set, and $\lambda \in [0, 1]$. The purpose of mixup is to teach the convolutional neural network to map a linear combination of inputs to a linear combination of outputs. Hence, by training neural networks on such combinations of inputs, mixup encourages the linear behaviors of the models, improves the generalization errors of cutting-edge network architectures, alleviates the memorization of corrupt labels, and increases sensitivity to adversarial examples \citep{mixup}.

\begin{figure}[h]
\begin{center}
\includegraphics[width = \textwidth]{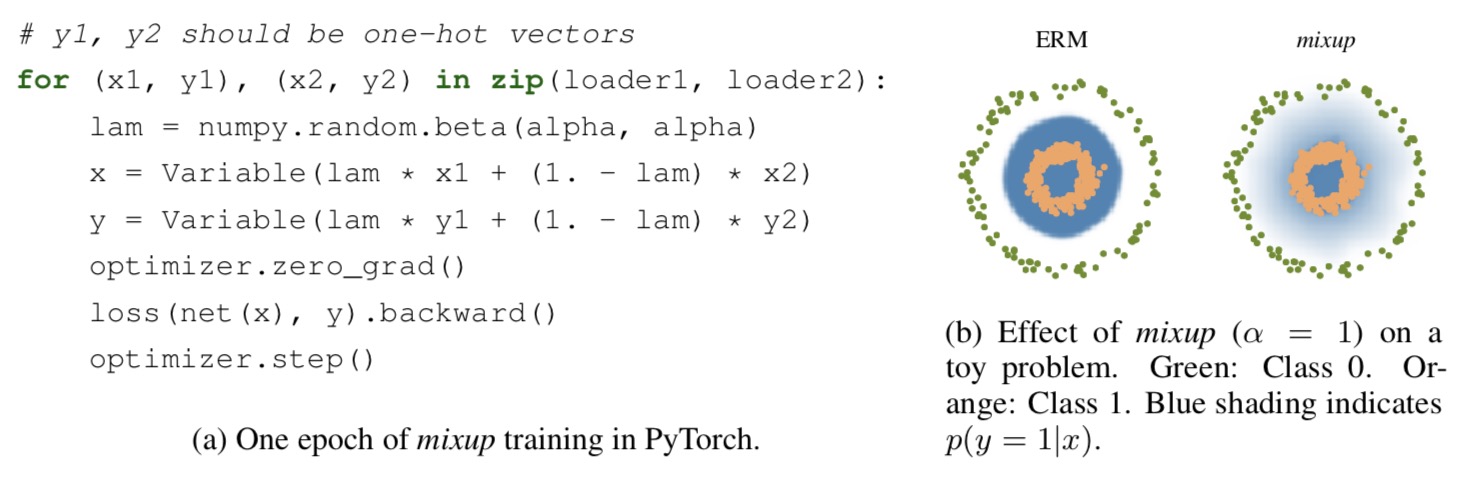}
\end{center}
\caption{Illustration of mixup, from \citet{mixup}.}
\label{fig:mixup}
\end{figure}

In addition, we considered the cutout approach proposed in DeVries \& Taylor (2017). This method randomly masks out contiguous square regions of the input images to CNNs, thereby augmenting the training data with partially occluded versions of existing samples. In particular, cutout applies a fixed-size zero-mask to a random portion of each input during each epoch. In this way, the technique encourages the neural network to take more of the full image context into account instead of relying on specific visual features. Thus, cutout improves robustness and yields higher accuracy levels \citep{cutout}.

\begin{figure}[h]
\begin{center}
\includegraphics[width = 0.8\textwidth]{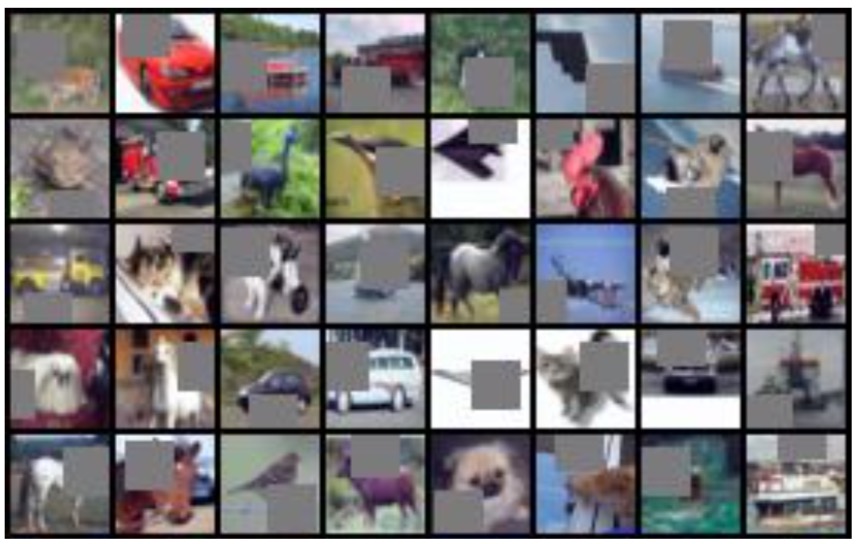}
\end{center}
\caption{Cutout applied to images from the CIFAR 10 dataset, from \citet{cutout}.}
\label{fig:mixup}
\end{figure}

\subsection{Pruning}
We made use of soft filter pruning, a technique that simultaneously prunes unimportant filters and trains the model \citep{he2018soft}. By simultaneously pruning and training the model, this method allows almost all of the layers to be pruned at once rather than pursuing the greedy layer by layer pruning methodology. More specifically, this algorithm can be broken down into four main parts.

First, we select filters from each layer that will be pruned. The number of filters chosen is equal to the product of the number of total filters and a pruning hyperparameter that is used for every layer. In particular, we set the pruning rate hyperparameter to be $0.9$. Within a layer, a filter is selected with probability proportional to its $l_p$ norm. After selection, these filters are set to $0$, i.e. are pruned. Then, the network is trained for one epoch, where the pruned filters may be updated to be non-zero due to backpropagation. These previous steps are repeated until convergence. At convergence, the zero filters are removed from the network. This generates the compact model.

\begin{figure}[h]
\begin{center}
\includegraphics[width = \textwidth]{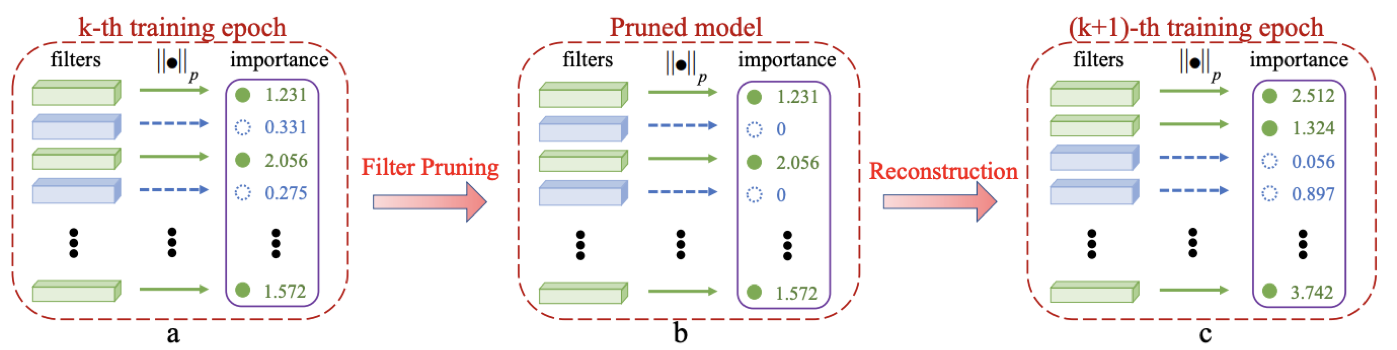}
\end{center}
\caption{Steps of soft filter pruning, from \citet{he2018soft}.}
\label{fig:softfilterpruning}
\end{figure}

\section{Implementation}

We leveraged PyTorch and PyTorch Lightning for implementing and training the models \citep{pytorch, lightning}. Additionally, we harnessed Weights $\&$ Biases for saving and visualizing the results \citep{wandb}. 

Link to our project:

\url{https://github.com/taivu1998/ResNet-Regularization-Pruning}

\section{Experiments}
Pruning methods and regularization techniques have been used independently to improve the computational efficiency of ResNet models. However, there has not been much work done in determining the computational complexity of the models when both techniques are implemented together. We show that on the ResNet architecture and the CIFAR 10 dataset, pruning methods and regularization techniques implemented together work in conjunction, rather than interfere with each other, to reduce computational inefficiency more than when either is implemented independently.

\subsection{Experiment Design}
To investigate the impact of using pruning and regularization together, we considered four different types of networks:
\begin{enumerate}
    \item Control network: This is the standard ResNet architecture, with no additions. We will be testing this model in order to have a point of comparison with the regularized (2) and pruning (3) networks. Using this, we will be able to isolate the effect that these methods have on the control network individually to gain better insight on our results from the combination.
    \item Network with regularization: This network has either mixup or cutout on top of the control network, with all other parameters fixed. Using this network, we will be able to directly compare to the combined (4) network, in order to determine whether achieves an equivalent ($\pm 5\%$) accuracy with a lower FLOP count.
    \item Network with pruning: This is the control network with soft filter pruning. We will be able to use this network to also compare to the combined (4) network, and whether we can achieve better results.
    \item Network with both pruning and regularization: This network has both regularization (mixup or cutout) and soft filter pruning combined. The goal of this network is to reduce the number of FLOPs from the regularization (2) network, without decreasing accuracy by more than $5\%$.
\end{enumerate}
For our control networks, we implemented ResNet 20, ResNet 32, ResNet 44, ResNet 56, and ResNet 110 based on Facebook's models \citep{resnet}. We considered two regularization techniques, mixup and cutout, and trained ten different types of networks with regularization: either mixup or cutout applied on each of the five control models \citep{cutout, mixup}. For pruning, we harnessed soft filter pruning \citep{he2018soft}. Each of the five control models then had pruning added. Lastly, we constructed our fourth class of networks by combining soft filter pruning and each of mixup and cutout to each of the control models.

\subsection{Dataset}

We used the CIFAR 10 dataset, which has $60,000$ color images \citep{cifar10}. Each of them is an RGB image of size $32 \times 32 \times 3$. These images are equally split into ten different classes: airplane, automobile, bird, cat, deer, dog, frog, horse, ship, and truck. From these examples, $10,000$ of them are allocated as test images. The goal of these networks is to correctly allocate as many of these pictures into the correct class of the ten. The accuracy score is determined by dividing the number of correct classifications over the total number of images in the test set.

\subsection{Evaluation}
When implementing pruning and regularization techniques to our model, we say that our model does not affect accuracy if the accuracy is at most $5\%$ lower than that of the control network with the same size. There are several other variables that we controlled for, including the number of epochs each model was run for, the learning rate, and the dataset we were testing on. Specifically, we trained our networks on the CIFAR 10 data for 200 epochs with a learning rate of $0.1$. By having these control variables, we are able to minimize the differences between our models and ensure that our results are the effect of changes in our independent variable.

Our dependent variable is the computational efficiency of our ResNet-based models on the CIFAR 10 dataset. We determined the computational efficiency of each by calculating the number of floating-point operations (FLOPs). FLOPs is the number of floating point operations carried out by a model, and is a proxy for measuring its computational efficiency. It solely depends on the architecture of the model, with no regards to how the model is trained. We calculated FLOPs manually based on the channel and width of each layer of the network using the method from \citet{he2018soft}. FLOPs allow us to precisely determine the computational cost of a model in a consistent way. The goal of our research is to reduce the cost of running deep neural networks. By measuring FLOPs and working on reducing them, we think that the results of our research can be used to limit the cost of running DNNs in real industry use cases. Throughout our paper we refer to the FLOP count of a model as its cost. 

Our independent variable is which technique is implemented on the standard ResNet model \citep{resnet}. More specifically, we had ResNet models with the following methods implemented on top of them: mixup, cutout, soft filter pruning, mixup and soft filter pruning combined, and cutout and soft filter pruning combined. For each of these cases we also altered the number of layers in the ResNet architecture (20, 32, 44, 56, 110) to get a broader view of how these techniques affect the different network sizes. We show consistent results across all network sizes to help make our results more robust and more widely applicable to more use cases. This helps minimize the risks related to the effects of certain network sizes on the metrics that we are measuring.

The goal of our evaluation is to illustrate that we can reduce the computational complexity of artificial neural networks during the inference phase without any significant loss of accuracy. In particular, for each type of the ResNet structure, we show that a combination of regularization and pruning can decrease its number of FLOPs while keeping its accuracy level comparable. To accomplish this, we calculated both the FLOP count and the accuracy score of every combination of regularization and pruning. 

We expect that for each control model (i.e. ResNet 20, 32, 44, 56, 110), adding regularization will improve accuracy and leave the FLOP count constant, while adding pruning will decrease both FLOPs and accuracy. The goal of the research is to show that the fourth class of networks reduces the number of FLOPs without sacrificing accuracy. We tested four different sizes of each of the five network types (20, 32, 44, 56, 110), in order to get a curve of five different points of FLOPs and accuracy. Using this, we can show how the curves move on the graph with network changes. Our goal is for the (4) curve to have both lower FLOP counts than (2) and higher accuracies than (1) and (3) across all network sizes.

\section{Results}

\begin{table}[]
\caption{Results: Accuracy and MegaFLOPs of all methods and network sizes.}
\label{table: results}
\begin{center}
\begin{tabular}{l|l|c|c}
\textbf{Method} &\textbf{Model} & \textbf{MegaFLOPs} & \textbf{Accuracy ($\%$)} \\
\hline
Control & ResNet 20  & $\hspace{5pt}40.55$   & $91.63$    \\
Control & ResNet 32  & $\hspace{5pt}68.86$   & $92.11$    \\
Control & ResNet 44  & $\hspace{5pt}97.17$   & $92.54$    \\
Control & ResNet 56  & $125.49$  & $92.49$    \\
Control & ResNet 110 & $252.89$  & $92.58$   \\
\hline
Mixup & ResNet 20  & $\hspace{5pt}40.55$     & $92.67$    \\
Mixup & ResNet 32  & $\hspace{5pt}68.86$     & $93.39$    \\
Mixup & ResNet 44  & $\hspace{5pt}97.17$     & $94.16$    \\
Mixup & ResNet 56  & $125.49$    & $94.15$    \\
Mixup & ResNet 110 & $252.89$    & $94.71$   \\
\hline
Cutout & ResNet 20  & $\hspace{5pt}40.55$    & $93.00$   \\
Cutout & ResNet 32  & $\hspace{5pt}68.86$    & $93.15$   \\
Cutout & ResNet 44  & $\hspace{5pt}97.17$    & $93.10$   \\
Cutout & ResNet 56  & $125.49$   & $93.01$   \\
Cutout & ResNet 110 & $252.89$   & $92.76$  \\
\hline
Pruning & ResNet 20  & $\hspace{5pt}34.37$   & $91.09$   \\
Pruning & ResNet 32  & $\hspace{5pt}58.58$   & $91.40$   \\
Pruning & ResNet 44  & $\hspace{5pt}82.78$   & $92.23$   \\
Pruning & ResNet 56  & $106.99$  & $91.42$   \\
Pruning & ResNet 110 & $215.92$  & $91.88$   \\
\hline
Mixup \& Pruning & ResNet 20  & $\hspace{5pt}34.37$   & $91.94$  \\
Mixup \& Pruning & ResNet 32  & $\hspace{5pt}58.58$   & $93.06$   \\
Mixup \& Pruning & ResNet 44  & $\hspace{5pt}82.78$   & $93.12$   \\
Mixup \& Pruning & ResNet 56  & $106.99$  & $93.80$  \\
Mixup \& Pruning & ResNet 110 & $215.92$  & $93.04$  \\
\hline
Cutout \& Pruning & ResNet 20  & $\hspace{5pt}34.37$   & $92.87$   \\
Cutout \& Pruning & ResNet 32  & $\hspace{5pt}58.58$   & $93.28$   \\
Cutout \& Pruning & ResNet 44  & $\hspace{5pt}82.78$   & $94.12$   \\
Cutout \& Pruning & ResNet 56  & $106.99$  & $94.52$   \\
Cutout \& Pruning & ResNet 110 & $215.92$  & $94.57$  
\end{tabular}
\end{center}
\end{table}

\subsection{Control}
\begin{figure}[h]
\begin{center}
\includegraphics[width = \textwidth]{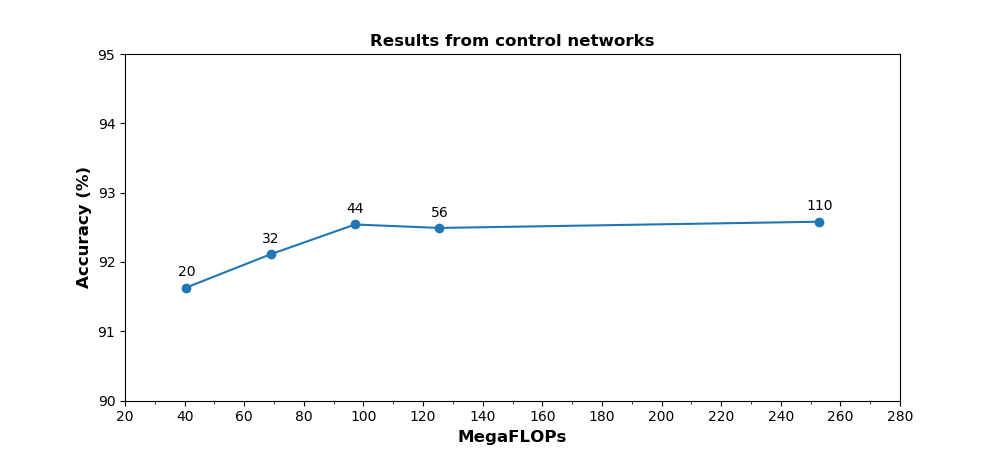}
\caption{Results from control networks. Five different labeled network sizes of ResNet with corresponding MegaFLOPs and accuracy when run on CIFAR 10.}
\label{fig:control}
\end{center}
\end{figure}

We ran all five sizes of the control network, which are recorded in Table \ref{table: results} and visualized in Figure \ref{fig:control}. We can see that these models have decent performances on the CIFAR 10 dataset, with the predictive accuracies ranging from $91.63 \%$ to $92.58 \%$. These statistics are similar to those in \citet{he2016deep}.

When it comes to computational cost, the number of FLOPs increases as the size of the network rises, with $40.6$ megaFLOPs for ResNet 20 and $252.9$ megaFLOPs for ResNet 110. This result is completely reasonable because larger networks have more weights and require more computations when processing a certain input. We have verified that the FLOP counts we obtained on our control models (i.e. with no techniques implemented) are consistent with the calculations from \citet{he2018soft} in addition to those from another paper by \citet{li2016pruning}.

Meanwhile, it is evident that the accuracy level tends to go up with increasing network sizes, but the increase is no more than $1 \%$. This means that it is not worthwhile to utilize a network larger than ResNet 32, because beyond that, the cost rises dramatically without a considerable accuracy increase. For the rest of the analysis, we have continued running on all network sizes, in order to observe whether our results apply in other cases.

\subsection{Regularization}
\begin{figure}
\begin{center}
\includegraphics[width = \textwidth]{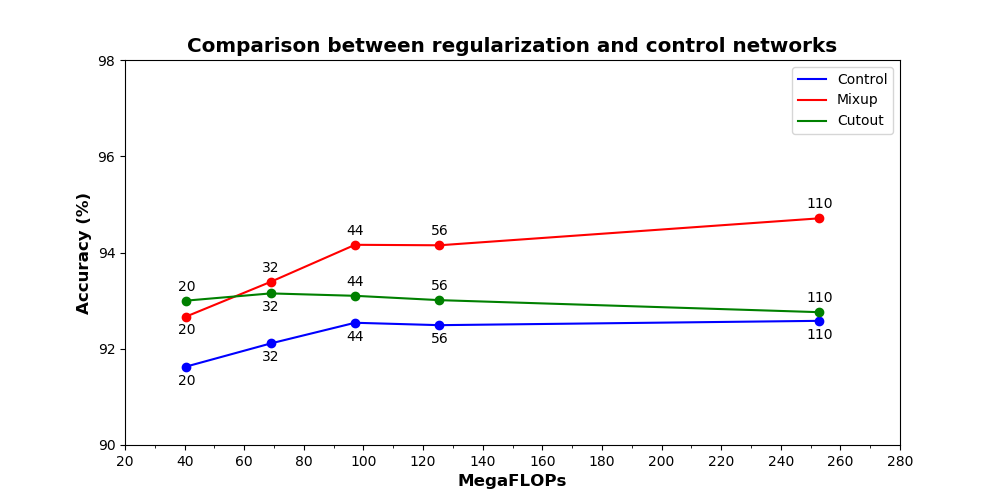}
\caption{Results from regularized networks compared with control networks. Five different labeled network sizes of ResNet with corresponding MegaFLOPs and accuracy when run on CIFAR 10.}
\label{fig:regularization}
\end{center}
\end{figure}

When implementing mixup and cutout on all five of our control models (Table \ref{table: results} and Figure \ref{fig:regularization}), there are consistent increases in accuracy level as shown in Table \ref{table: regularization}. In fact, all cutout models undergo accuracy increases of $0.18 - 1.37\%$. Likewise, mixup seems to have a greater effect, as it raises the accuracy scores of all five networks by $1.04 - 2.13\%$.  These figures illustrate that both mixup and cutout are highly effective at improving the generalization error and enhancing the accuracy level of the ResNet architecture. While the goal of this paper is to reduce FLOPs while keeping accuracy consistent, this accuracy increase could help overcome decreases that could result from combining this with pruning. 

Additionally, for mixup, the improvement tends to become more significant with increasing network sizes. Specifically, ResNet 110 experiences a $2.13\%$ accuracy increase by adding mixup. Therefore, the mixup models show an upward trend in accuracy levels as we go from ResNet 20 to ResNet 110. However, with regards to cutout, larger networks demonstrate less remarkable effects, with the accuracy increases of $1.37\%$ and $0.18\%$ for ResNet 20 and ResNet 110 respectively. At the same time, all five cutout models have comparable performances with no more than $0.4\%$ differences in accuracy.

Furthermore, the addition of regularization does not affect the FLOP counts of the models. This fact is to be expected, as regularization just changes the values of, not the number of parameters in a neural network.

\begin{table}
\caption{Increase in accuracy of regularized vs. control networks.}
\label{table: regularization}
\begin{center}
\begin{tabular}{l|l|c}
\textbf{Method} &\textbf{Model} & \textbf{Increase in Accuracy ($\%$)} \\
\hline
Mixup & ResNet 20  & $1.04$    \\
Mixup & ResNet 32  & $1.28$    \\
Mixup & ResNet 44  & $1.62$    \\
Mixup & ResNet 56  & $1.66$    \\
Mixup & ResNet 110 & $2.13$   \\
\hline
Cutout & ResNet 20  & $1.37$    \\
Cutout & ResNet 32  & $1.04$    \\
Cutout & ResNet 44  & $0.56$    \\
Cutout & ResNet 56  & $0.52$    \\
Cutout & ResNet 110 & $0.18$   
\end{tabular}
\end{center}
\end{table}

\subsection{Pruning}
\begin{figure}[h]
\begin{center}
\includegraphics[width = \textwidth]{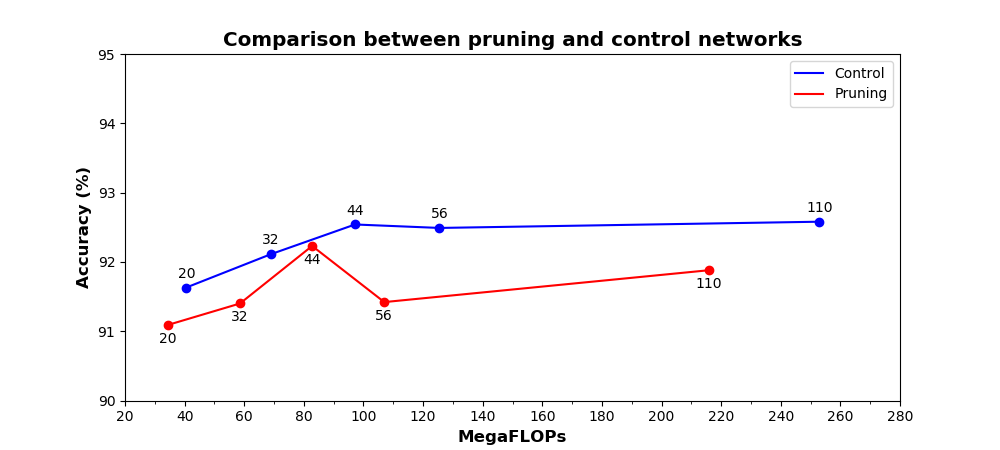}
\caption{Results from pruning networks compared with control networks. Five different labeled network sizes of ResNet with corresponding MegaFLOPs and accuracy when run on CIFAR 10.}
\label{fig:pruning}
\end{center}
\end{figure}

As shown in Figure \ref{fig:pruning} and Table \ref{table: pruning}, soft filter pruning reduces the accuracy scores by $1.07\%$ in the worst case across all five network sizes, which is much less than the threshold that we set of a maximum $5\%$ decrease. This is a minimal accuracy decrease that may be recovered by combining this technique with regularization. More interestingly, we can see that the effect of pruning becomes more significant as we go from ResNet 20 to ResNet 44, before the accuracy decreases. One possible explanation is that for ResNet 44, filter pruning sharply reduces the numbers of parameters and prevents overfitting. Larger networks like ResNet 56 and ResNet 110 still have a vast quantity of weights after being pruned, so they are still prone to overfitting and generalize less competently to unseen data.

With regards to FLOP counts, all 5 model sizes experience roughly $15\%$ declines (Table \ref{table: pruning}). These are sharp decreases in the computational costs that do not significantly affect the accuracies. Thus, soft filter pruning shows promise for a method that could help achieve the goal of creating DNNs that perform more efficient inference.

\begin{table}[]
\caption{Pruning effect on accuracy and FLOPs.}
\label{table: pruning}
\begin{center}
\begin{tabular}{l|c|c}
\textbf{Model} & \textbf{Decrease in Accuracy ($\%$)} & \textbf{Decrease in FLOPs ($\%$)} \\
\hline
ResNet 20 & $0.16$ & $15.2$  \\
ResNet 32 & $0.71$ & $14.9$   \\
ResNet 44 & $0.31$ & $14.8$   \\
ResNet 56 & $1.07$ & $14.7$   \\
ResNet 110 & $0.70$ & $14.6$ \\
\end{tabular}
\end{center}
\end{table}

\subsection{Combined}
\begin{figure}[h]
\begin{center}
\includegraphics[width = 0.93\textwidth]{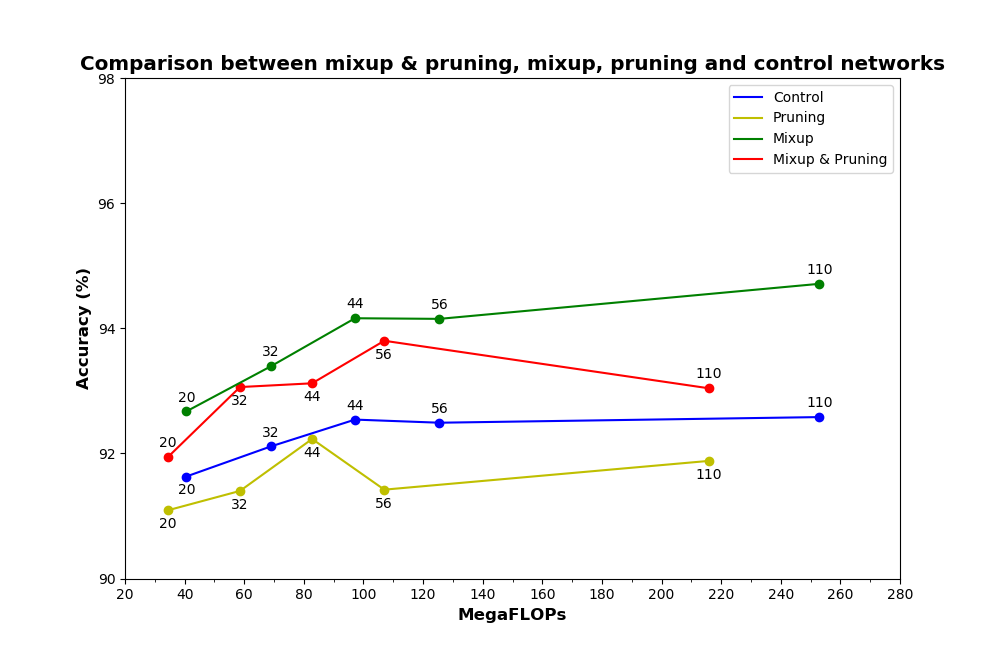}
\caption{Results from combined methods (mixup and pruning) compared to mixup, pruning and control individually. Five different labeled network sizes of ResNet with corresponding MegaFLOPs and accuracy when run on CIFAR 10.}
\label{fig:combined_mixup}
\end{center}
\end{figure}

\begin{figure}[h]
\begin{center}
\includegraphics[width = 0.93\textwidth]{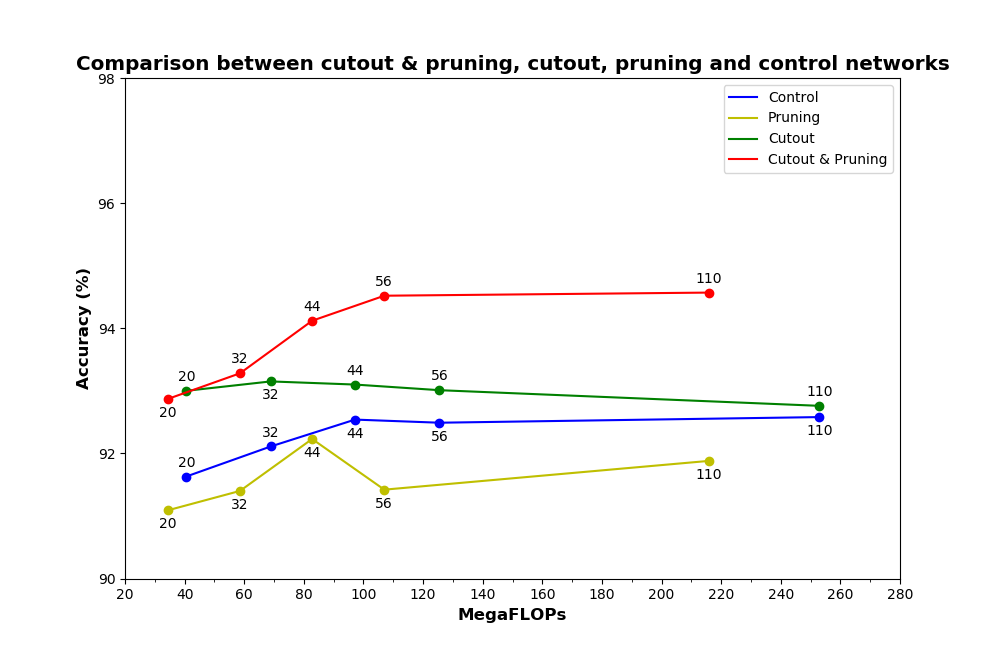}
\caption{Results from combined methods (cutout and pruning) compared to cutout, pruning and control individually. Five different labeled network sizes of ResNet with corresponding MegaFLOPs and accuracy when run on CIFAR 10.}
\label{fig:combined_cutout}
\end{center}
\end{figure}

The results of adding both soft filter pruning and regularization to the baseline models are illustrated in Table \ref{table:combined}, Figure \ref{fig:combined_mixup}, and Figure \ref{fig:combined_cutout}. We expected the predictive accuracies of the combined models to increase back to the range of just the regularization networks. 

As for the models with mixup and pruning implemented, their accuracy scores are consistently higher than those of the pruning networks and the control networks, with $0.31 - 2.38\%$ differences. These accuracy levels are also slightly lower than the figures for the mixup models, with decreases of no more than $1.67\%$ (less than the threshold of $5\%$).

In terms of the models with cutout and implemented, there results are even more notable. Indeed, there accuracy scores also exceed those of the pruning and control models by about $1.17 - 3.10\%$. More remarkably, their accuracy levels tend to surpass the figures for the cutout models, with rises of more than $1.81\%$ for ResNet 32, 44, 44, 56, and 110. In addition, the effect of combining cutout and pruning becomes increasingly significant for larger networks, which is demonstrated by an upward trend in the accuracy levels of these combined models.

When it comes to the computational costs, all ten networks have the same FLOP counts as the corresponding pruning models do. This is to be expected because regularization does not affect the costs. In other words, they show approximately $15\%$ reductions in the numbers of FLOPs with trivial impacts on accuracy.

Overall, based on the aforementioned statistics, we learn that the combination of regularization and pruning techniques results in compressed models that achieve lower computational costs and higher accuracy scores. Therefore, our proposal that the two techniques can effectively combine without conflicting is validated by the data.

\begin{table}
\caption{Increase in accuracy of combined vs. control networks.}
\label{table:combined}
\begin{center}
\begin{tabular}{l|l|c|c}
\textbf{Method} &\textbf{Model} & \textbf{Increase in Accuracy (\%)} & \textbf{Decrease in FLOPs (\%)} \\
\hline
Mixup \& Pruning & ResNet 20  &  $0.31$  & $15.2$ \\
Mixup \& Pruning & ResNet 32  &  $0.95$  & $14.9$ \\
Mixup \& Pruning & ResNet 44  &  $0.58$  & $14.8$ \\
Mixup \& Pruning & ResNet 56  &  $1.31$  & $14.7$ \\
Mixup \& Pruning & ResNet 110 &  $0.46$  & $14.6$ \\
\hline
Cutout \& Pruning & ResNet 20  & $1.24$  & $15.2$ \\
Cutout \& Pruning & ResNet 32  & $1.17$  & $14.9$ \\
Cutout \& Pruning & ResNet 44  & $1.58$  & $14.8$ \\
Cutout \& Pruning & ResNet 56  & $2.03$  & $14.7$ \\
Cutout \& Pruning & ResNet 110 & $1.99$  & $14.6$
\end{tabular}
\end{center}
\end{table}

\section{Discussion}
For each given network size, the accuracy of the combined model is greater than the figures for the control network and the network with pruning. However, the number of FLOPs of the combined model is the same as that of the model with pruning and much less than that of the control network. Hence, pruning is a complement to regularization, and combining them produces a superior performance with a good tradeoff between accuracy and computational cost.

It is interesting to note that the FLOP count decreases by roughly $15\%$ when pruning is implemented. This large decrease is particularly surprising given that the pruning rate was set to $0.9$, i.e. $10\%$ of the parameters were pruned from each layer every time. Hence, we can see that soft filter pruning is particularly effective for reducing the computational complexity of ResNet during inference time.

In the future it would be interesting to conduct additional experiments with varied pruning rates such as $0.8$ and $0.7$. Perhaps a smaller compression rate will result in a model with fewer learnable parameters, which is less suspectible to overfitting. Hence, such a model may even have a higher accuracy level with a lower FLOP count.

In addition, we will apply a combination of regularization and pruning to larger models sizes like ResNet 18 and ResNet 34 and larger datasets like ImageNet. With more training data and larger networks that contain much more parameters, the effects of mixup, cutout, and soft filter pruning may become more significant. It is also interesting to run experiments with other popular network architectures like AlexNet, VGGNet, GoogleNet, UNet, and DenseNet and observe whether our conclusion applies to these models.

Furthermore, we are currently evaluating our models almost solely on their numbers of FLOPs, with the criteria that their accuracies must be above the specified threshold. We would also like to construct a metric that is better able to take into account percent changes in both cost and accuracy. This would allow us to develop more nuanced insights into the results of the combined models compared to the models with just regularization or pruning.

\section{Conclusion}
In this paper we proposed that combining pruning and regularization methods would perform better than the baseline and either technique individually. In particular, we define the performance by the reduction in the cost, or the FLOP count, of the model from the control network of the same size, such that the change in accuracy is minimal ($<5\%$ decrease). 

In comparison with the control network, models with regularization implemented increase the accuracies, while those with soft filter pruning decrease the costs. Combining both the regularization and pruning techniques results in models with higher accuracies than those with just pruning. Meanwhile, compared to models with only regularization, these combined models have both significantly lower costs (about $15\%$) and slightly lower (or even higher) accuracies (up to $2\%$). Consequently, these results substantiate our hypothesis. 

Further research could look into different regularization and pruning techniques and other ways of combining pruning and regularization that can boost the performance of artificial neural networks during inference time.
\nocite{*}
\newpage 

\bibliographystyle{iclr2020_conference}
\bibliography{iclr2020_conference}

\end{document}